%% file: bare_conf.tex
\documentclass[a4paper,conference]{IEEEtran}
\usepackage[utf8]{inputenc}
\usepackage{multirow}
\usepackage{amsmath}
\usepackage{lipsum}
\usepackage[english]{babel}
\usepackage{subfiles}
\usepackage{float}
\usepackage{subcaption}
\usepackage{tabularx}
\usepackage{pgfplots}
\usepackage{verbatim}
\usepackage{bm}
\usepackage{paralist}
\let\llncssubparagraph\subparagraph
\let\subparagraph\paragraph
\usepackage[compact]{titlesec}
\let\subparagraph\llncssubparagraph
\usepackage[font={scriptsize}]{caption}
\usetikzlibrary{calc}
\usepackage{kpfonts}

\xdefinecolor{tucol1}{rgb}{1.0 0.5 0.05}
\xdefinecolor{tucol2}{rgb}{0.52,0.72,0.10}
\xdefinecolor{tucol3}{rgb}{0.0 0.0 0.8}
\xdefinecolor{tucol4}{rgb}{0.8 0.8 0}
\xdefinecolor{tucolv}{rgb}{0.1 0.35 0.17}
\xdefinecolor{tucol5}{rgb}{0.8 0 0.8}
\xdefinecolor{tucol6}{rgb}{0 0.8 0.8}
\xdefinecolor{tucol7}{rgb}{0.7 0.8 0.1}
\xdefinecolor{tucol8}{rgb}{0.2 0.7 0.34}
\xdefinecolor{tucol9}{rgb}{0.7 0.8 0.3}
\xdefinecolor{tucol10}{rgb}{0.5 0.5 0.5}
\xdefinecolor{tucol11}{rgb}{0.9 0.9 0.3}
\xdefinecolor{blue}{rgb}{0.4 0.5 1}
\xdefinecolor{softgreen}{rgb}{0.49,0.50,0.47}
\titlespacing{\section}{0pt}{2ex}{1ex}
\titlespacing{\subsection}{0pt}{1ex}{0ex}
\titlespacing{\subsubsection}{0pt}{0.5ex}{0ex}

\usepackage{graphicx}
\graphicspath{{Images/}}

\usepackage{enumitem}
\setlist[itemize]{align=parleft,left=0pt..1em}

\ifCLASSINFOpdf
\else
\fi
\begin{document}
%
\title{Video-based Pose-Estimation Data as Source for Transfer Learning in Human Activity Recognition}

\author{\IEEEauthorblockN{Shrutarv Awasthi$^*$, Fernando Moya Rueda$^{*}$, and Gernot A. Fink}
\IEEEauthorblockA{Pattern Recognition in Embedded Systems Group\\ 
TU Dortmund University, Dortmund, Germany\\
\{shrutarv.awasthi,fernando.moya,gernot.fink\}@tu-dortmund.de}
}


%


\maketitle
\def\thefootnote{*}\footnotetext{These authors contributed equally to this work.}
\begin{abstract}
Human Activity Recognition (HAR) using on-body devices identifies specific human actions in unconstrained environments. HAR is challenging due to the inter and intra-variance of human movements; moreover, annotated datasets from on-body devices are scarce. This problem is mainly due to the difficulty of data creation, i.e., recording, expensive annotation, and lack of standard definitions of human activities. Previous works demonstrated that transfer learning is a good strategy for addressing scenarios with scarce data. However, the scarcity of annotated on-body device datasets remains. This paper proposes using datasets intended for human-pose estimation as a source for transfer learning; specifically, it deploys sequences of annotated pixel coordinates of human joints from video datasets for HAR and human pose estimation. We pre-train a deep architecture on four benchmark video-based source datasets. Finally, an evaluation is carried out on three on-body device datasets improving HAR performance. 
\end{abstract}


%
\IEEEpeerreviewmaketitle

\input{chapters/01_introduction}

\input{chapters/02_relatedwork}

\input{chapters/03_method}

\input{chapters/04_experiments}

\input{chapters/05_conclusions}


\section*{Acknowledgment}
\footnotesize{This work was supported by Deutsche Forschungsgemeinschaft (DFG) in the context of the project Fi799/10-2 and HO2463/14-2 ''Transfer Learning for Human Activity Recognition in Logistics'', also a part of the project 45KI02B021 ''Silicon Economy Logistics Ecosystem '' funded by the German Federal Ministry of Transport and Digital Infrastructure, and Olga Robertina Moya de Suárez,  (\textborn 1953-\textdagger 2021).}




\bibliographystyle{IEEEtranS}
\bibliography{literatur}



\end{document}

%% file: chapters/01_introduction.tex
\section{Introduction}
    
    Human activity recognition (HAR) concerns classifying activities of human movements. HAR is nowadays essential for applications in Industry 4.0, ambient-assisting living, health support, and smart-homes, e.g., activities of daily living (ADLs) \cite{chikhaoui2018cnn,hammerla2016_DCRMHARUW,liu2016human,niemann2020lara}. HAR methods mostly use signals from videos, marker-based motion-capturing systems (marker-based Mocap), or on-body devices. The latter comprise different sensors, e.g., accelerometers, gyroscopes, and magnetometers. On-body devices are also non-invasive, unaffected by occlusion, and do not portray personal identities. Therefore, these devices are suitable for HAR as activities can be tracked and performed in the natural environment. Nevertheless, the high inter-class and intra-class variation of human activities make HAR a challenging task. Besides, HAR datasets suffer from the class imbalance problem. 

    Deep learning-based approaches have been used successfully for solving multi-channel time-series HAR \cite{chikhaoui2018cnn,hammerla2016_DCRMHARUW,Moya_Rueda2020,ramasamy2018recent,yang2015deep}. Deep networks in HAR are layered end-to-end architectures that learn simple to abstract features of human movements. They can learn non-linear and temporal relations of basic, complex, and highly dynamic-human movements from raw time-series data. 
    They capture the local dependencies of multi-channel time-series data, which are also translation-invariant in time \cite{chen2020}. Deep networks learn more discriminative features of human actions in contrast to statistical pattern-recognition approaches \cite{yang2015deep,zeng2014convolutional}.
    
    However, the performance of multi-channel time-series HAR using deep networks has not shown a significant increase in prediction accuracy as in other fields, such as image and video classification \cite{Moya_Rueda2020,reining2019human}. Considering HAR as a supervised problem, scarcity of annotated HAR data is the primary concern \cite{demrozi2020human,kim2009human}. Supervised learning methods, e.g. deep neural networks, require a large amount of annotated data. The annotation process demands enormous resources. Additionally, annotation reliability varies based on human errors or on unclear and non-elaborated annotation protocols \cite{avsar_2021,feldhorst2016motion,kim2009human,reining2020annotation}.

    Transfer learning can alleviate the problem of scarcity of annotated data. However, it can be hindered by the enormous variation of recording settings, e.g., different recording rates, sensor resolutions, device position, or intrinsic device characteristics. Unlike transfer learning in computer vision tasks where source datasets are large, e.g., Imagenet dataset \cite{Krizhevsky2012-ICD}, source datasets for multi-channel time-series HAR remain scarce. Accordingly, we propose using datasets intended for video-based HAR and human pose estimation as a source for HAR. Specifically, we use the annotations of pixel-coordinate as sequences of human joints. This proposal relates to hybrid approaches of human-pose estimation and HAR in video-based tasks \cite{luvizon20182d,usman2019multi}.
    
    This paper is structured as follows. Sec.~\ref{chap:rw} discusses the deep learning-based methods and transfer learning for HAR. Sec.~\ref{chap:method} presents the method for implementing the transfer learning using human poses and their derivatives. Sec.~\ref{sec:dataset} introduces the three source and three target datasets used in this work. Sec.~\ref{sec:exp} presents and discusses the results obtained from various experiments. Finally, Sec.~\ref{sec:conclusion} draws the inferences.

%% file: chapters/02_relatedwork.tex
\section{Related Work}
\label{chap:rw}

    
    Statistical pattern recognition methods are common for analysing human movements from measurements from on-body devices. A standard pipeline involves pre-processing, segmentation, hand-crafted feature extraction, and classification. Pre-processing is necessary due to sensor characteristics, sampling rate, and noise. Segmentation is commonly carried out employing a sliding window approach on the sensor measurements along the time axis. Statistical features are extracted either from the time domain or frequency domain \cite{bulling2014tutorial,feldhorst2016motion,laguna2011dynamic,olszewski2001generalized,reining2019human,twomey2018comprehensive}. These features are aggregated using PCA, LDA, or KDA for dimensionality reduction and finally used for training a set of parameters of a classifier. The classifier assigns a class label to an unknown sequence using its extracted and aggregated features \cite{lara2012survey,reining2019human}.

    Nowadays, deep learning methods are relevant for solving HAR problems. Deep architectures holistically combine feature extraction and classification, e.g., the temporal convolutional neural networks (tCNNs) for HAR \cite{grzeszick2017deep,twomey2018comprehensive}. \cite{cruciani2020feature,duffner_2014,grzeszick2017deep,moya2018_CNNHARBWS,ronao2015deep,yang2015deep} employed convolution and pooling operations along the temporal dimension to capture local temporal dependencies. Besides, the convolutional filters are shared among all the sensors. The authors in \cite{ordonez2016deep} proposed a deep architecture for HAR, which combines convolution and recurrent layers. This DeepConvLSTM consists of four convolutional layers and three LSTM layers. 
    The authors observed that the DeepConvLSTM offers better performance when identifying the start and end of activities. Also, DeepConvLSTM improved the classification performance when compared to results reported by \cite{yang2015deep}, using a four-layered tCNN. 
    The authors in \cite{grzeszick2017deep,moya2018_CNNHARBWS} proposed a tCNN based architecture for HAR in an industrial setting. The architecture processes data from on-body devices in separate parallel blocks. Temporal convolutions are performed for each block. The temporal convolution layer of each on-body device shares the same weights. A fully connected layer computes an intermediate representation for each parallel block. A subsequent fully connected layer then fuses the intermediate representations. Joining the information of individual on-body devices later in the CNN architecture, i.e., late fusion, makes the IMU-CNN more powerful against slightly asynchronous and inherent characteristics of on-body devices.
    
    HAR can also be performed using human pose estimates. In \cite{luvizon20182d}, authors combined human pose estimations from images and sequences of RGB frames to create an end-to-end architecture for HAR from videos. The authors used a differentiable soft-argmax for pose estimation. This layer allowed an action recognition network to be stacked on top of the pose estimator, resulting in an end-to-end trainable network. In \cite{cippitelli2016human}, the authors proposed an activity recognition algorithm exploiting skeleton data extracted by RGBD sensors. Human joint-poses are recorded using a Microsoft Kinect. Subsequently, a multiclass Support Vector Machine classifies activities.
    
    Supervised deep learning methods require substantial labelled data to produce decent results. However, annotated on-body devices data is scarce as the annotation process is expensive, time-consuming, tedious, and requires domain expertise \cite{avsar_2021,chen2020,reining2020annotation}. In the context of deep networks, transfer learning gives an alternative to the problem of scarcity of annotated data \cite{Moya_Rueda2020,ordonez2016deep_conv}. Transfer learning uses a network trained on a source task to initialise a network on a related target one. In \cite{ordonez2016deep_conv}, authors characterised the feasibility, benefits, and drawbacks of performing transfer learning in three scenarios: subjects within a dataset, different datasets, on-body device locations, and sensor type. They created a network with three convolutional layers alternated with max-pooling layers, followed by an LSTM and a softmax layer. The authors considered windows of $9.7$ sec. They varied the number of transferred layers. For the first case, the authors concluded that the filters of the lower layers are more generic, thus, transferable. The authors inferred that performance varies when the source and target datasets are different for the second case; the type of activities in the source domain negatively affects the transfer learning performance. Finally, there was significant degradation in performance when transferring between modalities and locations irrespective of the number of transferred layers.
    
    In \cite{chikhaoui2018cnn}, authors investigated the transfer learning performance on three scenarios: across subjects of different ages using the same on-body device on the same location, different positions of on-body devices, and different sampling rates and sensor types. The authors inferred that transfer learning between any on-body device placement is possible. In addition, lower layers capture generic features independent of the sampling rate. In \cite{Moya_Rueda2020}, authors introduced a method for transfer learning from human joint-poses from a Mocap system as a source and inertial measurements obtained from on-body devices as the target dataset. They used a tCNN \cite{yang2015deep} and an IMU-CNN \cite{grzeszick2017deep,moya2018_CNNHARBWS}, trained using sequences of human joint-poses or their derivatives. They used a sliding window of $1$ sec. The authors performed transfer learning across three target domains with different activities, numbers of on-body devices, and recording rates. The authors inferred that the performance improved for both networks on three target datasets. Furthermore, results are valid even when fine-tuning with a proportion of the datasets.
    
    Transfer learning has been carried out among datasets from on-body devices. These datasets remain, however, scarce or limited sized. Besides, they contain annotations of task-related activities. Therefore, exploiting other data sources might be interesting for transferring purposes in HAR.

%% file: chapters/03_method.tex
\section{Human Poses for HAR}
\label{chap:method}
    Considering the conclusions in \cite{Moya_Rueda2020}, sequences of human poses serve as a source for transfer learning for multi-channel time-series HAR purposes. We propose to extend it by considering sequences of human poses from data of different purposes. We use \textbf{annotations} of pixel-coordinate sequences of human joints from video data intended for video pose estimation. These datasets contain annotated recordings from different scenarios with a broad range of human activities in the wild. Thus, we seek to squeeze the usability of these datasets for multi-channel time-series HAR. These human-pose annotations from videos can be considered multi-channel time-series of human movements. We use the second derivative of a smooth piecewise spline interpolation of degree five on a small-time interval from sequences of human joint-poses for simulating on-body devices attached to the poses, a sort of synthetic data. In line with \cite{gurjar2018_LDRWSUWS,Moya_Rueda2020}, synthetic data could also be beneficial for transfer learning, improving performance on a limited-sized datasets.  Fig.~\ref{fig:networks} shows the proposed method.
    
    \input{figures/network}
    
\subsection{Temporal Convolutional neural network}
    \label{section:tcnn}
    
    The temporal convolutional neural network (\textit{tCNN}) and the \textit{tCNN-IMU} are used as feature extractors and classifiers, based on \cite{grzeszick2017deep,Moya_Rueda2020,moya2018_CNNHARBWS,niemann2020lara,yang2015deep}. 
    The \textit{tCNN-IMU} \cite{grzeszick2017deep} consists of five branches, each with four temporal-convolution layers ($c1,c2,c3,c4$) having $64$ filters of size $[5,1]$. The five branches are assigned measurements of the four limbs and the torso, i.e., left-arm (LA), right-arm (RA), left-leg (LL), right-leg (RL), and head-neck or torso (N). If the measurements for one or more branches are missing, then those branches are removed. A fully-convolutional layer computes a representation per block. The five representations are concatenated and fed to a fully-convolutional layer. Finally, a softmax layer is present at the end to classify human activities. The \textit{tCNN} can be considered as an architecture with a single block. Both architectures successively process a sequence of human activity windows of dimensions $[W,D]$. 
    Each architecture is trained using batch gradient descent with the RMS-Prop update rule with a momentum of $0.9$, weight decay of $5\times10^{-4}$, and three different learning rates $[10^{-3},10^{-4},10^{-5}]$. The learning rate for every experiment is selected according to the results on the validation set. Dropout is applied after the first and second fully-connected layers. The categorical cross-entropy loss function is used. Orthonormal initialization is deployed for initializing the weights. A data augmentation technique adds Gaussian noise $[\mu =0,\sigma =0.01]$ to input data. 
    

%% file: figures/network.tex
\begin{figure}[!ht]

    \centering

    \begin{tikzpicture} [x=0.4cm,y=0.4cm][scale=0.2]
        \tikzstyle{node1}=[text=black, font=\tiny \bfseries];
        \tikzstyle{node2}=[text=black, font=\tiny \bfseries];
        \tikzstyle{node3}=[text=black, font=\tiny];
        \tikzstyle{node4}=[text=black, font=\tiny];
        \tikzstyle{arrow1} = [line width=0.5]
        \tikzstyle{circle1}=[circle,draw=black, minimum size=0.1cm, line width=0.2mm, inner sep=0pt]
        \tikzstyle{circle2}=[circle,draw=black, minimum size=0.05cm, line width=0.1mm, inner sep=0pt, fill=black]
       
       \node (label) at (0.0 - 15.5, 2.2 + -0.2)[]{
                \includegraphics[width= 0.05\textwidth]{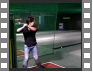}
                };
       
       \node (label) at (0.0 - 15.5, 2.2 + 0.1)[]{
                \includegraphics[width= 0.05\textwidth]{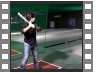}
                };
       
        \node (label) at (0.0 - 15.5, 2.2 + 0.6)[]{
                \includegraphics[width= 0.05\textwidth]{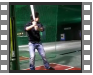}
                };
        \node (label) at (0.0 - 15.5, 2.2 + 1.0)[]{
                \includegraphics[width= 0.05\textwidth]{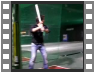}
                };
        \node (label) at (0.0 - 15.5, 2.2 + 1.4)[]{
                \includegraphics[width= 0.05\textwidth]{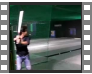}
                };
        \node [node1] at (0.0-14.5,1.5 - 0.9){Annotated Pose Data};
         \draw [-latex](0-13.5,0+3.2) -- (0.0 -11.5, 0.0 + 3.2);
        \node [node1] at (0.0-9.3,0.9 + 3.8){Human Pose};
        \node [node1] at (0.0-16,0.9 + 3.8){Video};
        \node [node1] at (0.0-12.5,0.9 + 4.2){Source Dataset};
         \node (label) at (0.0 - 9, 0.0 + 0.7 + 2)[]{
                \includegraphics[width= 0.072\textwidth]{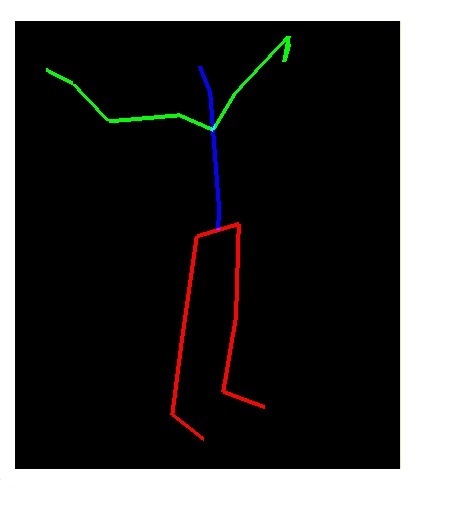}
              };
        
        \node (label) at (0.0 -14, 0.0 - 1.7)[]{
                \includegraphics[width= 0.14\textwidth]{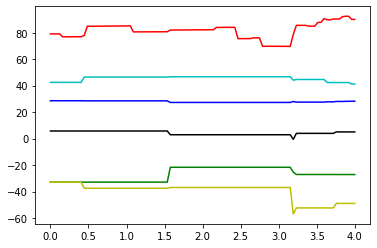}
              }; 
        
        \node (label) at (0.0 -9.5, 0.0 - 1.7)[]{
                \includegraphics[width= 0.14\textwidth]{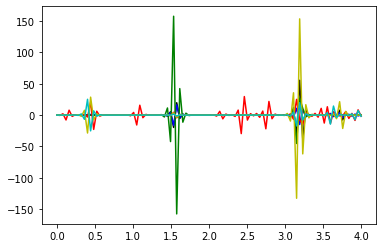}
              }; 
           
        \node (label) at (0.0 -1 , 0.0 + 0.7)[]{
                \includegraphics[width= 0.23\textwidth]{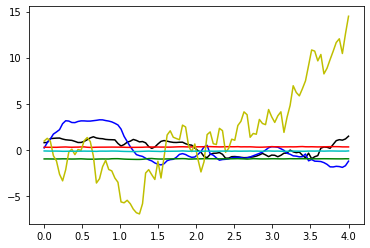}
              }; 
        \draw [rounded corners,line width=0.05mm, opacity=0.8] (0.0-17.2,0.0-4 ) rectangle +(11,9.5);

        \node [node1] at (0.0-9.1,0.6){Synthetic Data};
        \node [node3] at (0.0+2,-2.8){Time};
        
              
        \node [node3] at (0.0-12.5,-1.5){D};
        \node [node3] at (0.0-9.2,-3.8){Time};
        \node [node3] at (0.0-5.9,+0.8){D};
        \node [node3] at (0.0-1.2,0.9 -5.7){Convolution layers};
        \node [node3] at (0.0-11.8,0.9 -5.7){Convolution layers};
        
        \node [node3] at (0.0-12,0.0-7.2){$C=64$};
        \node [node3] at (0.0-8.8,0 - 4.8){FC};
        \node [node3] at (0.0+5.9-13,0.9 - 5.7){Softmax};
        \node [node3] at (0.0+6.4-13.5,0+1.7-7){Walk};
        \node [node3] at (0.0+6.4-13.7,0+1.3-7){Sit};     
        \node [node3] at (0.0+6.4-13.6,0+0.9-7){Run}; 
        \node [circle2] at (0.0+6.4-13.5,0.5-7){};
        \node [circle2] at (0.0+6.4-13.5,0.2-7){};
        \node [circle2] at (0.0+6.4-13.5,0-7.1){};
        \draw [rounded corners,line width=0.05mm, opacity=0.8] (0.0-14,0.0 - 7.5) rectangle +(4.2,2.5);
        \draw [rounded corners,line width=0.05mm, opacity=0.8] (0.0-9.6,0.0 - 7.5) rectangle +(1.5,2.5);
        \draw [rounded corners,line width=0.05mm, opacity=0.8] (0.0-7.8,0.0 - 7.5) rectangle +(1.5,2.5);
            \draw (1.1-10.6,0-6.2) -- (2.5 -10.6, 0.0 -6.2);
         \draw (1.1,0-6.2) -- (2.5 , 0.0 -6.2);
        
        
        
 \draw [line width=0.05mm,fill=tucol11, scale=0.7, opacity=1](1.2 + 2.8-22.5,2.2-10)--(0.4 + 2.8-22.5,2.2-10)--(0 + 2.8-22.5,1.9-10)--(0.8 + 2.8-22.5,1.9-10)--(1.2 + 2.8-22.5,2.2-10)--(1.2 + 2.8-22.5,0.3-10)--(0.8 + 2.8-22.5,0-10)--(0 + 2.8-22.5,0-10)--(0 + 2.8-22.5,1.9-10)--(0.8 + 2.8-22.5,1.9-10)--cycle; 
         \draw [line width=0.05mm,scale=1](-13.16,4.35-10)--(-13.16,3-10);
      \draw [line width=0.05mm,fill=tucol11, scale=0.7, opacity=1](1.2 + 4.15-22.5,2.2-10)--(0.4 + 4.15-22.5,2.2-10)--(0 + 4.15-22.5,1.9-10)--(0.8 + 4.15-22.5,1.9-10)--(1.2 + 4.15-22.5,2.2-10)--(1.2 + 4.15-22.5,0.3-10)--(0.8 + 4.15-22.5,0-10)--(0 + 4.15-22.5,0-10)--(0 + 4.15-22.5,1.9-10)--(0.8 + 4.15-22.5,1.9-10)--cycle; 
        \draw [line width=0.05mm,scale=1](-12.24,4.35-10)--(-12.24,3-10);
     \draw [line width=0.05mm,fill=tucol11, scale=0.7, opacity=1](1.2 + 5.5-22.5,2.2-10)--(0.4 + 5.5-22.5,2.2-10)--(0 + 5.5-22.5,1.9-10)--(0.8 + 5.5-22.5,1.9-10)--(1.2 + 5.5-22.5,2.2-10)--(1.2 + 5.5-22.5,0.3-10)--(0.8 + 5.5-22.5,0-10)--(0 + 5.5-22.5,0-10)--(0 + 5.5-22.5,1.9-10)--(0.8 + 5.5-22.5,1.9-10)--cycle; 
       \draw [line width=0.05mm,scale=1](-11.32,4.35-10)--(-11.32,3-10);
        \draw [line width=0.05mm,fill=tucol11, scale=0.7, opacity=1](1.2 + 6.85-22.5,2.2-10)--(0.4 + 6.85-22.5,2.2-10)--(0 + 6.85-22.5,1.9-10)--(0.8 + 6.85-22.5,1.9-10)--(1.2 + 6.85-22.5,2.2-10)--(1.2 + 6.85-22.5,0.3-10)--(0.8 + 6.85-22.5,0-10)--(0 + 6.85-22.5,0-10)--(0 + 6.85-22.5,1.9-10)--(0.8 + 6.85-22.5,1.9-10)--cycle; 
       \draw [line width=0.05mm,scale=1](-10.42,4.35-10)--(-10.42,3-10);
       
       \draw [fill=tucol11, line width=0.05mm, opacity=1] (1.5-10.7,-5.5) rectangle +(0.2,-1.5);
        \draw [fill=tucol11, line width=0.05mm, opacity=1] (2-10.7,-5.5) rectangle +(0.2,-1.5);

        
        


        
        \node [node1] at (0-0.5,4.7){Real On-body data};
        \node [node1] at (0-0.5,4.2){(Target Dataset)};
              
         \draw [-latex,line width=1pt](0-0.6,0 - 2.8) -- (0-0.6, 0.0 - 4.2);
         
       \draw [line width=1pt,scale=1](-15,3-7)--(-15,2-8);
       \draw [-latex,line width=1pt](0 -15,0-6) -- (0.0 -14, 0.0 - 6);
       \node [node1] at (0-16.1,0-5){Annotated};
       \node [node1] at (0-16,0-5.5){Pose};
       \node [node1] at (0-16,0-6){(or)};
       \node [node1] at (0-15.6,0-6.5){Synthetic Data};
       \node [node1] at (0-5,0-5.2){Transferring};
       \node [node1] at (0-4.9,0-5.6){Convolutional};
       \node [node1] at (0-5,0-6.3){Layers};
       \draw [-latex,line width=1pt](0-6.1,0-6) -- (-3.5, 0.0 -6);
       
        \node [node3] at (0.0 - 1,0.0-7.2){$C=64$};
        \node [node3] at (0.0+1.9,0 - 4.7){FC};
        \node [node3] at (0.0+3.5,3-7.7){Softmax};
        \node [node3] at (0.0+3.5,1.7-7){Wave};
        \node [node3] at (0.0+3.5,1.3-7){Drink};     
        \node [node3] at (0.0+3.5,0.9-7.1){Jump}; 
        \node [circle2] at (0.0+3.5,0.5-7){};
        \node [circle2] at (0.0+3.5,0.2-7){};
        \node [circle2] at (0.0+3.5,0-7.1){};
        \draw [rounded corners,line width=0.05mm, opacity=0.8] (0.0-3.3,0.0 - 7.5) rectangle +(4.2,2.5);
        \draw [rounded corners,line width=0.05mm, opacity=0.8] (1.1,0.0 - 7.5) rectangle +(1.5,2.5);
        \draw [rounded corners,line width=0.05mm, opacity=0.8] (2.8,0.0 - 7.5) rectangle +(1.4,2.5);
        
\draw [line width=0.05mm,fill=tucol11, scale=0.7, opacity=1](1-4.1,2.2-10)--(0.2-4.1,2.2-10)--(-0.2-4.1,1.9-10)--(0.6-4.1,1.9-10)--(1-4.1,2.2-10)--(1-4.1,0.3-10)--(0.6-4.1,0-10)--(-0.2-4.1,0-10)--(-0.2-4.1,1.9-10)--(0.6-4.1,1.9-10)--cycle; 
        \draw [line width=0.05mm,scale=1](-2.45,4.35-10)--(-2.45,3-10);
       \draw [line width=0.05mm,fill=tucol11, scale=0.7, opacity=1](1-4.1+1.35,2.2-10)--(0.2-4.1+1.35,2.2-10)--(-0.2-4.1+1.35,1.9-10)--(0.6-4.1+1.35,1.9-10)--(1-4.1+1.35,2.2-10)--(1-4.1+1.35,0.3-10)--(0.6-4.1+1.35,0-10)--(-0.2-4.1+1.35,0-10)--(-0.2-4.1+1.35,1.9-10)--(0.6-4.1+1.35,1.9-10)--cycle; 
        \draw [line width=0.05mm,scale=1](-1.5,4.35-10)--(-1.5,3-10);
     \draw [line width=0.05mm,fill=blue, scale=0.7, opacity=1](1-4.1+2.7,2.2-10)--(0.2-4.1+2.7,2.2-10)--(-0.2-4.1+2.7,1.9-10)--(0.6-4.1+2.7,1.9-10)--(1-4.1+2.7,2.2-10)--(1-4.1+2.7,0.3-10)--(0.6-4.1+2.7,0-10)--(-0.2-4.1+2.7,0-10)--(-0.2-4.1+2.7,1.9-10)--(0.6-4.1+2.7,1.9-10)--cycle; 
       \draw [line width=0.05mm,scale=1](-0.53,4.35-10)--(-0.53,3-10);
       
      \draw [line width=0.05mm,fill=blue, scale=0.7, opacity=1](1-4.1+4.05,2.2-10)--(0.2-4.1+4.05,2.2-10)--(-0.2-4.1+4.05,1.9-10)--(0.6-4.1+4.05,1.9-10)--(1-4.1+4.05,2.2-10)--(1-4.1+4.05,0.3-10)--(0.6-4.1+4.05,0-10)--(-0.2-4.1+4.05,0-10)--(-0.2-4.1+4.05,1.9-10)--(0.6-4.1+4.05,1.9-10)--cycle; 
        \draw [line width=0.05mm,scale=1](+0.42,4.35-10)--(+0.42,3-10);
      
        \draw [fill=blue, line width=0.05mm, opacity=1] (1.5,-5.5) rectangle +(0.2,-1.5);
        \draw [fill=blue, line width=0.05mm, opacity=1] (2,-5.5) rectangle +(0.2,-1.5);

    \end{tikzpicture}

   \caption{\scriptsize{A tCNN is pre-trained on either pixel-coordinate sequences of joints or their derivatives for HAR. The convolutional layers from the pre-trained network are transferred to a tCNN being trained on three on-body devices datasets for HAR.
   \vspace{-5mm}
   }}
    \label{fig:networks}
\end{figure}
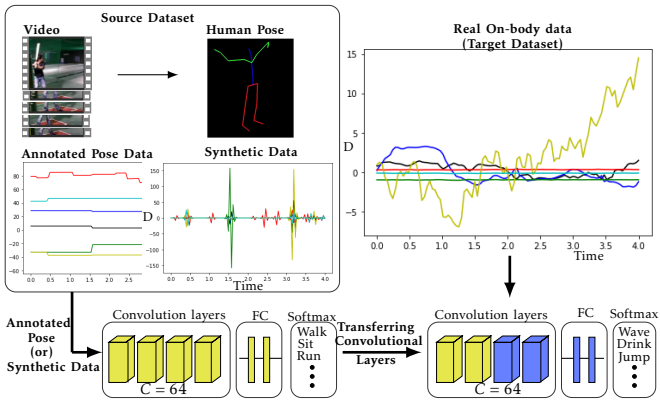


%% file: chapters/04_experiments.tex
\section{Datasets}
\label{sec:dataset}
    We use various source datasets $D_s$ containing human-pose annotations as the ground-truth annotations of video datasets for HAR and pose estimation, namely the JHMDB, CAD60, Penn Action and the NTU RGB+d benchmark datasets. These pose data comprises 2D (x,y) pixel coordinates or 3D (x,y,z) with an additional depth annotation. Annotations along each axis are treated as a separate channel. Target datasets $D_t$ comprise on-body devices recordings of human activities. These datasets are the Pamap2, Opportunity, and LARa. The $D_s$ are split into training ($70\%$), validation($15\%$), and testing($15\%$) sets, following \cite{ordonez2016deep}.
    

    \subsection{Source Datasets ($D_s$)}
    
        \textbf{JHMDB} \cite{Jhuang:ICCV:2013} contains $928$ video clips of different human actions in the wild. The dataset comprises $21$ activities. The person performing the activity in each frame is manually annotated with their $2D$ joint positions. 
        The joint positions correspond to LA, LL, RA, RL, and N. Every activity class contains $36-55$ video clips, and each clip contains $15-40$ frames. The video clips are recorded at a rate of $25$ Hz. The duration of activities ranges from $0.5-2$ sec. 
            
        \textbf{CAD-60} \cite{sung2014cornell} contains RGB-D video sequences of human activities, which are recorded using the Microsoft Kinect v1 \cite{6190806}. The dataset contains $[320\times 240]$ sized RGB-D motion sequences and skeletal information acquired at $30$ Hz. The skeletal information is composed of $15$ $3D$-joint positions per skeleton. The $3D$ coordinates are extracted from the depth data. The joint positions correspond to LA, LL, RA, RL, and N. The activities are performed by four subjects in five different constrained environments.
        The subjects perform $12$ activities. The total recorded time for all the activities is approximately $47$ min.
            
        \textbf{Penn Action} \cite{zhang2013actemes} consists of $2326$ videos, recorded at $50$ Hz, of different human actions in the wild. The annotations comprise $15$ activity classes, $13$ 2D human joint-poses in each video frame, and camera viewpoints. The dataset contains $13$ 2D positions of human joints, specifically from head, shoulders, elbows, wrists, hips, knees and ankles. The total recorded time for all the activities is approximately $55$ min.  
        
            
        \textbf{NTU RGB+D} \cite{shahroudy2016ntu} contains $60$ activity classes and $56,880$ video samples. The dataset contains RGB videos, depth map sequences, 3D skeletal data, and infrared (IR) videos for each sample. Videos are captured from $40$ different subjects, using Microsoft Kinect v2 at $30$ Hz. Skeletal data consists of $3D$ locations of $25$ major body joints for detected and tracked human bodies. The joint locations correspond to LA, LL, RA, RL, and N.
        

    \subsection{Target Datasets ($D_t$)}
        \textbf{Pamap2} \cite{reiss2012introducing} contains recordings of three on-body devices from nine subjects performing Activities of daily living. Here, the on-body devices are placed on the dominant ankle, dominant hand, and chest. Pamap2 dataset uses a recording rate of $100$ Hz. The subjects perform $18$ activities. The total recorded time for all the activities is approximately $324$ min. The
        validation set contains recordings from subject $5$, and testing from subject $6$.      
        
        \textbf{LARa} \cite{niemann_friedrich_2020_3862782,niemann2020lara} contains recordings of human poses and inertial measurements from $14$ subjects performing activities in the intralogistics. The dataset provides recordings of an Optical Marker-based Motion Capture (OMoCap), called LARa-MoCap, with a recording rate of $200$ Hz, and two sets of on-body devices, with a recording rate of $100$ Hz, called LARa-Mbientlab (LARa-Mb) and LARa-MotionMiners (LARa-MM). The LARa-MM considers only three on-body devices and on different locations than LARa-Mb. In LARa-MoCap, $3D$ poses of $22$ joints are present. LARa-Mb consists of $3D$ linear and angular acceleration measured from five on-body devices attached to LA, LL, RA, RL, and torso. Whereas, LARa-MM consists of measurements from three on-body devices; LA, RA, and N. The dataset is strongly unbalanced. The \textit{handling (centred)} class constitutes $60\%$ of the complete recordings. For LARa-MoCap, the validation and test set contain recordings from subjects $[5,11,12]$ and $[6,13,14]$ respectively. Similarly, for LARa-Mb and LARa-MM, the validation set consists of measurements from subjects $[11,12]$ and testing from subjects $[13,14]$.
            
        \textbf{Opportunity} \cite{chavarriaga2013opportunity} contains recordings of seven on-body devices from four subjects performing ADLs. Subjects perform the activities in a room simulating a studio flat with a deckchair, a kitchen, doors giving access to the outside, a coffee machine, a table, and a chair. The activities were recorded using seven on-body devices, twelve 3D-acceleration sensors, four 3D localization information, and object sensors, with a recording rate of $30$ Hz. The sensor measurements correspond to LA, LL, RA, RL, and N. The user's activities were annotated on different levels: 17 mid-level gesture classes (Opp-Gest) and five high-level locomotion (Opp-Loc) classes. The validation set contains the ADL3 recordings from subjects $[2,3]$, and testing ADL4 and ADL5 recordings from subjects $[2,3]$. Other recordings are a part of the training set.
        
        

\section{Experiments and Results}
\label{sec:exp}
    The \textit{IMU-tCNN} with $five$ branches, each per human limb, and a \textit{tCNN}, in Fig.~\ref{fig:networks}, are pre-trained on different $D_s$. The convolutional layers of the pre-trained networks \textit{IMU-tCNN}$_{D_s}$ and \textit{tCNN}$_{D_s}$ are transferred to a \textit{tCNN} to be trained on the $D_t$, denoted by \textit{tCNN}$_{D_s}^{D_t}$. The number of transferred convolutional layers ($N_{conv}$) and the $\%$ of $D_t$ for fine-tuning vary$^1$. The transferred convolution layers are not frozen$^2$. The non-transferred convolution and fully-connected layers of the \textit{tCNN}$_{D_s}^{D_t}$ are trained from scratch. HAR is a multi-class classification problem, and all datasets are highly unbalanced. Thus, the weighted $F1$ score ($wF1$) \cite{ordonez2016deep} is used as a performance metric. This metric weighs each activity class equally using precision and recall. The results presented in this work are the mean and standard deviation $(\mu\pm \sigma)$ from five runs, with a random seed of $42$. A permutation test is performed to evaluate the performance changes. The testing accuracy will be considered for this permutation test \cite{ojala2010permutation}. For Penn Action, it is not possible to estimate a joint corresponding to a pose estimate in the dataset. As \textit{IMU-tCNN} uses branches for processing measurement from specific joints, therefore Penn Action cannot be trained using \textit{IMU-tCNN}.  
    The values in bold in all the tables have the corresponding testing accuracy significantly higher than the baseline one, based on a permutation test.
    
    \def\thefootnote{1}\footnotetext{Implementation of the two architectures, pre-processing of datasets, training and testing are found in https://github.com/shrutarv/Create-Synthetic-IMU-data.}
    \def\thefootnote{2}\footnotetext{Freezing the transferred convolutional layers showed lesser performance: thus, they are not presented here, however they can be found in the github.}
    \def\thefootnote{3}\footnotetext{Considering the joint-pose annotations as multi-channel time-series data for HAR defers from the video-based HAR approaches intended for $D_s$.}
    

    \subsection{HAR from Source Datasets}
    
        Following the method in Fig.~\ref{fig:networks}, HAR will be considered using transfer learning from ground-truth annotations of joint poses to real on-body data. The JHMDB, CAD60, Penn Action, and NTU will be used as $D_s$. Concretely, we consider the sequences of joint-pose annotations in pixel coordinates as multi-channel time-series data for HAR. Besides, their second order-derivatives will also be deployed as a source for multi-channel time-series HAR$^3$. The derivatives will be called synthetic on-body devices. This consideration takes the advantage that the \textit{IMU-tCNN} and \textit{tCNN} process sequences per channel with late fusion and local temporal-neighbourhoods of sequences are likely correlated. 
        The joint poses sequences of $D_s$ are either up-sampled or down-sampled to match the frequency of the $D_t$. The JHMDB, CAD60, Penn, and NTU are up-sampled by factors of $[4,3,2,3]$ respectively for $D_t$=\textit{Pamap2} as target and by factors of $[1,1,0.5,1]$ for $D_t$=\textit{Opp}. We deploy a smooth piecewise spline interpolation of degree five on a small time-interval from a sequence of joint-pose annotations per channel; accordingly, the second order-derivative of the spline approximation is used as synthetic on-body devices. 
        A sliding window approach with $T=1$sec. $W$ depending on the sampling rate, and a stride of $s$ is used for segmenting sequences of human poses; specifically, the following parameters were used: a window size of ($W=25$) and a stride of $s=12$ for JHMDB$_{pose}$, a window size of ($W=30$) and a stride of $s=12$ for the CAD60$_{pose}$, a window size of ($W=50$) and a stride of $s=1$ for the Penn$_{pose}$ and a window size of ($W=30$) and a stride of $s=3$ for the NTU$_{pose}$. 
        
        
        Channels are normalized to zero-mean and unit deviation. Following the pre-processing protocol in \cite{niemann2020lara}, we normalize the joint-poses with respect to the torso for JHMDB and CAD60, head for Penn, and middle of the spine for NTU.
        We consider multi-channel time-series HAR using the tCNN on the four $D_s$. Table \ref{tab:ss} shows the classification performance in terms of $wF1[\%]$ on their testing sets. The $wF1[\%]$ on the JHMDB$_{synth.}$, CAD$_{synth.}$ and NTU$_{synth.}$, the synthetic on-body devices, decreases when compared to pose data. These results are due to the approximations involved in the creation of synthetic data. The HAR performance increases when up-sampling for JHMDB$_{synth.}$ and NTU$_{synth.}$ to $100$Hz and decreases when down-sampling for Penn$_{synth.}$ to $30$Hz. 
        
        \input{Tables/pose_datasets}
        
        We consider multi-channel time-series HAR using joint poses differently from the approaches in \cite{choutas2018potion,javidani2021learning,usman2019multi} as video-based HAR. Interestingly, for JHMDB$_{synth.}$, upsampled to $100$Hz, we obtain a mean classification accuracy of $90.53\%$, which is significantly higher than the one reported by \cite{choutas2018potion} ($85.5\%$), \cite{javidani2021learning} ($77.54\%$), and \cite{usman2019multi} ($83.1\%$). The predictions were unsegmented, and a majority voting was deployed before computing the $wF1[\%]$ so that a comparison would be fair. All three pieces of research focused on HAR from videos, using RGB frames, Optical Flow and Pose. The authors in \cite{choutas2018potion} introduced the PoTion representation to encode the motion of joint poses over a video clip as an input. They used the 3D ConvNet (I3D), proposed by \cite{carreira2017quo}. In \cite{usman2019multi}, the authors combined a two-streams CNN---RGB and Optical flow---with a prediction and the ground truth of human joints from videos. They utilized a fixed tree-like tensor for representing the joint poses. Besides, they deployed early, middle and late fusion of the streams. In \cite{javidani2021learning}, authors pre-trained a CNN to extract feature maps from optical-flow images. A $1D$-CNN is used to extract temporal information from these flow-feature maps. We consider HAR using the ground-truth of human poses from the $D_s$ as multi-channel time-series and not tree-like tensors. 
        

        
        \subsection{Transfer learning from $D_s$ to Pamap2\\}
        
            \input{results/Pamap2}
        \subsection{Transfer Learning from $D_s$ to LARa-Mb and LARa-MM\\}
       
            \input{results/LaraOB}
            \input{results/LaraMM}

        \subsection{Transfer Learning from $D_s$ to Opportunity\\}

\input{results/opportunity}

        \subsection{Comparison to state of the art}
            
            Tab.~\ref{tab:wf1} presents the performance of the three pre-trained architectures on the $D_t$. Compared with the Opportunity and Pamap2, pre-training the \textit{tCNN} with the four source datasets, with both the pose annotations or the synthetic on-body devices, significantly influences the HAR's performance on the $D_t$, especially when transferring the first convolutional layer. 
            The more sophisticated networks, \textit{B-LSTM} and \textit{tCNN-LSTM}, present the highest performance. Additional experiments using these types of networks are required. However, following the results in Tab.~\ref{tab:TL_S_Pamap2}-\ref{tab:TL_S_Loc}, the performance in all scenarios with the \textit{tCNN} significantly improves when considering a lesser quantity of training material for the target dataset ($10,30,50,75$)[\%], pre-training with the four source datasets with pose annotations, or the synthetic on-body devices.
       
            \begin{table}
                \centering
                \caption{\scriptsize{The $wF1\%$ of the best $tCNN_{D_s}^{D_t}$ on the four target datasets compared to similar works using the same architecture. The results from \cite{Moya_Rueda2020}, denoted with "*", show the performance of pre-training the network with a large MoCap dataset.}}
                \resizebox{\columnwidth}{!}{
                \begin{tabular}{c|c|c|c| c|c} 
                \hline
                \textbf{Dataset} & \textbf{LARa-Mb} &\textbf{LARa-MM} &\textbf{Pamap2} & \textbf{Opp (Loc.)} & \textbf{Opp (Ges)} \\
                \hline
                B-LSTM \cite{hammerla2016_DCRMHARUW} &- &- & -& - & 90.8\\ 
                tCNN-LSTM\cite{ordonez2016deep}& - &- & - & 89.5 & 91.5\\ 
                tCNN \cite{yang2015deep}& - &- & - & 86.5 & 93.9\\ 
                tCNN\cite{Moya_Rueda2020} & 75.75$\pm$0.4 & - & 87.04 $\pm$0.4 & 84.53 $\pm$0.2 & 88.20$\pm$0.4\\ 
                tCNN*\cite{Moya_Rueda2020} & 76.19$\pm$0.2 & - & 90.95 $\pm$0.4 & 88.43 $\pm$0.3 & 91.31$\pm$0.0\\ 
                \hline
                tCNN(replicated) & 75.06$\pm$0.01  & 65.71 $\pm$0.01 & 86.67 $\pm$0.06  & 85.67 $\pm$0.01 & 87.36 $\pm$0.05\\ [0.4ex] 
                tCNN$_{JHMDB}$ & 75.47 $\pm$0.01  & 65.71$\pm$0.02 & 89.70 $\pm$0.01  & 86.84$\pm$0.41 & 89.05$\pm$0.02\\
                tCNN$_{CAD60}$ & 74.94 $\pm$0.01  & 65.67$\pm$0.01 & 89.73 $\pm$0.00  & 86.87$\pm$0.02 & 89.11$\pm$0.51\\
                tCNN$_{Penn}$ & 75.24 $\pm$0.01  & 65.97$\pm$0.01 & 90.64 $\pm$0.85  & 86.70$\pm$0.01 & 88.96$\pm$0.23\\
                tCNN$_{NTU}$ & 75.90$\pm$0.01  & 64.79$\pm$0.01 & 91.23 $\pm$0.85  & 87.21$\pm$0.18 & 89.19$\pm$0.33\\
                \hline
                \multicolumn{6}{c}{\vspace{-10mm}}\\
                \end{tabular}
                }
                \label{tab:wf1}
                 
            \end{table}

%% file: Tables/pose_datasets.tex
\begin{table}[t]
    \caption{\scriptsize{The $wF1[\%]$ for HAR on the three $D_s$: JHMDB, CAD60 and Penn Action. The predictions on test set were unsegmented before computing the $wF1[\%]$. Multiple experiments were performed varying the stride $s$ for each $D_s$.}   {JHMDB} is not up-sampled for {JHMDB}$_{synth 30}$.}
    \resizebox{\columnwidth}{!}{
    \centering
    \begin{tabular}{p{24mm}|p{13.5mm}p{18mm}p{18mm}p{18mm}}
        \hline
        \textbf{Source Datasets $D_s$}  & \textbf{Poses} & \textbf{Synthetic OBD} & \textbf{Synthetic OBD}&\textbf{Synthetic OBD}\\ 
                                        & & & \textbf{[30Hz]}&\textbf{[100Hz]}\\ 
        \hline
        \hline
        \textbf{JHMDB[25Hz]}      & 50.90$\pm$0.05 & 26.58$\pm$0.06 & 26.58$\pm$0.06 & 85.68$\pm$0.81\\ 
        \textbf{CAD60[30Hz]}      & 75.75$\pm$0.02 & 50.11$\pm$0.06 &  50.11$\pm$0.06 & 57.16$\pm$0.06\\ 
        \textbf{Penn Action[50Hz]} & 88.05$\pm$0.27 & 95.21$\pm$0.14 & 67.14$\pm$0.32 & 96.00$\pm$0.08\\ 
        \textbf{NTU[30z]} & 30.07$\pm$0.15 & 6.70$\pm$1.35 & 6.70$\pm$1.35 & 36.59$\pm$0.49\\ 
        \hline
        \multicolumn{5}{c}{\vspace{-6mm}}\\
    \end{tabular}
    }
    \label{tab:ss}
\end{table}

%% file: Video-based Pose-Estimation Data as Source for Transfer Learning in Human Activity Recognition ICPR2022_arxiv/results/PAMAP2.tex
Tab.~\ref{tab:TL_S_Pamap2} shows the performance of the networks \textit{tCNN}$^{D_t}$ and \textit{tCNN-IMU}$^{D_t}$ on the Pamap2 dataset as target. The $\textit{tCNN}_{D_s}^{Pamap2}$ is pre-trained using the $D_s$: JHMDB, CAD60, Penn Action and NTU datasets. The performance on the Pamap2 dataset significantly increases when pre-training with the JHMDB. The \textit{tCNN}$_{JHMDB}^{Pamap2^{c1}}$ and \textit{tCNN}$_{JHMDB}^{Pamap2^{c1,c2}}$ significantly improve the HAR performance. These results suggest that the features learnt by lower layers of a \textit{tCNN} are rather generic, thus, transferable. Besides, transferring deeper layers affects the performance drastically, as deeper layers learn data-specific features. This finding can be concluded as the activity classes in CAD60 and Pamap2 are dissimilar. Considering a limited amount of dataset for training, i.e., $[10, 30, 50,75]\%$, the synthetic datasets from the three $D_s$ show a positive influence.


\input{Tables/pamap2_cnn}

Table~\ref{tab:TL_PR} shows the precision and recall for each activity class of the Pamap2 and the shared ones in the JHMDB datasets. The \textit{tCNN}$_{JHMDB}^{Pamap2}$ using the network with the $c1$ pretrained on the synthetic on-body devices of the JHMDB and finetuned on the Pamap2. For comparing the performance of the networks, the harmonic mean (HM) of precision and recall are computed for each activity---the highest HMs are highlighted in bold. The activities \textit{Climb Stairs}, \textit{Sit}, \textit{Run}, \textit{Walk}, and \textit{Stand} are common between the Pamap2 and JHMDB. The performance of these shared activities increase. Activities that are only in Pamap2, but are semantically near to those in JHMDB, also showed a boost in performance, e.g.,  
\textit{Climb Stairs}$_{JHMDB}$-\textit{Desc. Stairs}$_{Pamap2}$, 
\textit{Stand}$_{JHMDB}$-\textit{Ironing}$_{Pamap2}$, as person performs the activity standing and the activity is picking centred.



\input{Tables/pamap2_activities}

%% file: Tables/pamap2_cnn.tex
\begin{table}
    \centering
    \caption{\scriptsize{Mean $wF1[\%]$ of the \textit{tCNN}$_{D_s}^{Pamap2}$ and \textit{tCNN-IMU}$_{D_s}^{Pamap2}$ using the joint poses and the synthetic data. The $N_{conv}$ changes from $c_1$ to $c_{1,2,3,4}$ keeping $100\%$ of the $D_t$. Subsequently, the $N_{conv}$ corresponding to highest $wF1[\%]$ is fixed and $[10, 30, 50, 75]\%$ of the $D_t$ are deployed for fine-tuning. The \textit{tCNN}$^{Pamap2}$ and the \textit{tCNN-IMU}$^{Pamap2}$, trained on the training set of the Pamap2, denote the baseline. Std $wF1[\%]$ lies around $0.01$.}}
    \label{tab:TL_S_Pamap2}
    \resizebox{\columnwidth}{!}{
    \begin{tabular}{p{10mm} c c|c c|c c|c c|c}
        \hline
        \multicolumn{10}{c}{\bm{$tCNN_{D_s}^{Pamap2}$}} \\
        [0.2em]
        \textbf{Transf.} & \multicolumn{2}{c}{D$_s$=JHMDB} & \multicolumn{2}{c}{D$_s$=CAD60} & \multicolumn{2}{c}{D$_s$=Penn} & \multicolumn{2}{c}{D$_s$=NTU RGB+D} & \textbf{Baseline} \\ 
        [0.1em]
        \textbf{layers} &  \textbf{Synth}  & \textbf{Pose}  & \textbf{Synth}  & \textbf{Pose}  & \textbf{Synth} & \textbf{Pose}  & \textbf{Synth} & \textbf{Pose} &  \\ 
        [0.1em]
        \hline
        \hline
        $c_1$ & \textbf{89.70}   & \textbf{88.37} & \textbf{88.70}  & \textbf{89.73}  & \textbf{89.89}  & \textbf{88.30} & \textbf{90.78} & \textbf{89.20}  &  \multirow{4}{*}{86.67}      \\
        [0.1em]
        $c_{1,2}$         & 87.72   & \textbf{88.15} & 87.85  & \textbf{88.24} & \textbf{90.10} & {87.91}  & \textbf{90.36} & \textbf{89.66} &       \\  
        [0.1em]
        $c_{1-3}$      & 86.51 & \textbf{88.12}  & 87.23  & 87.48 & \textbf{90.46}  & 87.56 & \textbf{89.16} & \textbf{90.73} & \\  
        [0.1em]
        $c_{1-4}$    & {84.96}   & \textbf{88.13}  & 87.77  & 87.89  & \textbf{90.64}  & 87.10 &   \textbf{88.94} & \textbf{91.23} &   \\ 
        [0.1em]
        \hline
        \hline
        75\bm{$\%D_t$}   & \textbf{89.64} & \textbf{89.60}         & {88.16}  & {88.17}    & {86.46 } & 86.81  & \textbf{88.73} & {88.02} & 88.17       \\  
        [0.1em]
        50\bm{$\%D_t$}    & \textbf{89.63}    & {86.93} & {85.76}  & \textbf{89.03} & {83.91 } & {83.30}   & 85.95 & 85.15 &  85.96      \\  
        [0.1em]
        30\bm{$\%D_t$}    & \textbf{58.70}   & \textbf{49.00}   & \textbf{51.98}  & \textbf{50.27}        &    \textbf{74.14} & \textbf{73.04} & \textbf{81.98} & \textbf{70.71} & 44.47      \\
        [0.1em]
        10\bm{$\%D_t$}    & \textbf{50.97}   & \textbf{46.75}   & \textbf{47.26}  & \textbf{45.93}        &    \textbf{50.20} & \textbf{49.20}  & \textbf{43.93} & \textbf{51.36} & 40.96       \\
        [0.1em]
        \hline
        \multicolumn{8}{c}{\vspace{-1mm}}\\
    \end{tabular}
    }
    \resizebox{\columnwidth}{!}{
    \begin{tabular}{p{10mm} c c|c c|c c|c}
        \hline
        \multicolumn{8}{c}{\bm{$tCNN-IMU_{D_s}^{Pamap2}$}}  \\
        [0.2em]
        \textbf{Transf.} & \multicolumn{2}{c}{D$_s$=JHMDB} & \multicolumn{2}{c}{D$_s$=CAD60} & \multicolumn{2}{c}{D$_s$=NTU RGB+D} &\textbf{Baseline} \\ 
        \textbf{layers} &  \textbf{Synth}  & \textbf{Pose}  & \textbf{Synth}  & \textbf{Pose}  & \textbf{Synth}  &\textbf{Pose} & \textbf{\%wF1}  \\ 
        \hline
        \hline
        $c_1$ & 87.72   & 87.34 & 87.68   & \textbf{88.56}    & 87.24 & 86.23  & \multirow{4}{*}{86.91}      \\  
        $c_{1,2}$         & 87.52   & 86.54 & 86.73        & \textbf{88.12}   &   87.75 & 86.42  &   \\  
        $c_{1-3}$      & 85.35 & 86.38       & 85.84      & 87.67     &   \textbf{88.09} & 86.93  &  \\  
        $c_{1-4}$    & 80.22   & 86.34  & 85.93 & 86.81     & \textbf{88.52} & 87.01 &   \\ 
        \hline
        \hline
        75\bm{$\%D_t$}   & 86.25   & 86.23 & 86.14  & 85.84   & 88.41 & 86.52  & 88.06  \\  
        50\bm{$\%D_t$}    & \textbf{85.38}  & \textbf{85.65} & 83.63  & 83.49  & \textbf{85.93} & \textbf{84.91}  & 83.04  \\  
        30\bm{$\%D_t$}    & \textbf{67.71} & \textbf{70.01}  & \textbf{64.43} & 60.29 & \textbf{66.96} &  \textbf{73.68}  & 60.35 \\ 
        10\bm{$\%D_t$}    & 40.20   & 40.20  & 40.20 & 40.20 & \textbf{51.30}   & \textbf{59.59}  & 40.20       \\ 
        \hline
        \multicolumn{6}{c}{\vspace{-7mm}}\\
    \end{tabular}
    }
\end{table}

%% file: Tables/pamap2_activities.tex
\begin{table}
    \centering
    \caption{\scriptsize{Precision$[\%]$ and recall$[\%]$ values per activity of the JHMDB and Pamap2 datasets. The \textit{tCNN}$^{Pamap2}$ is considered as the baseline. The values in bold are selected based on the higher harmonic mean of precision and recall.}}
    \resizebox{\columnwidth}{!}{
    \begin{tabular}{c cccccc}
        \hline
        \multirow{2}{*}{Pamap2 Act.} &
        \multicolumn{2}{c}{tCNN$_{JHMDB_{synth.}}$} &
        \multicolumn{2}{c}{tCNN$^{Pamap2}$} &
        \multicolumn{2}{c}{tCNN$_{JHMDB_{synth.}^{c1}}^{Pamap2}$} \\
        & \text{Prec.}\% & \text{Rec.}\% & \text{Prec.}\% & \text{Rec.}\% & \text{Prec.}\% & \text{Rec.}\%  \\
        \hline
        \hline
        \text{Climb stairs}&97.87 &86.79 & 73.28 & 95.01 &\bf{90.34} & \bf{95.90} \\
        \text{Run}&65.08 &85.42 & 100 & 92.26 & \bf{100} & \bf{92.91} \\
        \text{Sit}&100 &89.13 & 94.27 & 94.32 & \bf{95.38} & \bf{94.07} \\
        \text{Stand}&94.23&100& 87.53 & 28.65 & \bf{86.46} &\bf{40.71} \\
        \text{Walk}&96.00 &88.89 & 92.65 & 97.86 & \bf{98.06} &\bf{99.47} \\
        \text{Rope Jump}&-&-&\bf{83.33} &\bf{100} & 77.10 & 100 \\
        \text{Lying}&-&-&98.79  &97.21  & {100} & {96.76} \\
        \text{Cycling}&-&-&100 & 96.23 & {100} & {96.55} \\
        \text{Nordic Walk}&-&-& \bf{100} & \bf{94.53} & 90.34 & 95.90 \\
        \text{Desc. Stairs}&-&-&80.47  &84.74  & \bf{86.51} & \bf{87.80} \\
        \text{Vacuuming}&-&-&98.26 & 74.14  & \bf{96.01} & \bf{91.73} \\
        \text{Ironing}&-&-&67.62  &100 & \bf{72.04} & \bf{99.12}\\
        \hline
        \multicolumn{6}{c}{\vspace{-6mm}}\\
    \end{tabular}}
    \label{tab:TL_PR}
\end{table}

%% file: results/LaraOB.tex
\label{subsection:TL_S_L}
\begin{table}
    \centering
    \caption {\scriptsize{Mean $wF1[\%]$ of the \textit{tCNN}$_{D_s}^{LARa-Mb}$ and \textit{tCNN-IMU}$_{D_s}^{LARa-Mb}$ using the joint poses and the synthetic data. The $N_{conv}$ changes from $c_1$ to $c_{1,2,3,4}$ keeping $100\%$ of the $D_t$. Subsequently, the $N_{conv}$ corresponding to highest $wF1[\%]$ is fixed and $[10, 30, 50, 75]\%$ of the $D_t$ are deployed for fine-tuning. The \textit{tCNN}$^{LARa-Mb}$, trained on the training set of the LARa-OB, denotes the baseline. Std $wF1[\%]$ lies around $0.01$.}}
    \label{tab:TL_S_LARaOB}
    \resizebox{\columnwidth}{!}{
    \begin{tabular}{p{10mm} c c|c c|c c|c c|c}
        \hline
        \multicolumn{10}{c}{\bm{$tCNN_{D_s}^{LARa-Mb}$}} \\
        [0.2em]
        \textbf{Transf.} & \multicolumn{2}{c}{D$_s$=JHMDB} & \multicolumn{2}{c}{D$_s$=CAD60} & \multicolumn{2}{c}{D$_s$=Penn} & \multicolumn{2}{c}{D$_s$=NTU RGB+D} & \textbf{Baseline} \\ 
        [0.1em]
        \textbf{layers} &  \textbf{Synth}  & \textbf{Pose}  & \textbf{Synth}  & \textbf{Pose}  & \textbf{Synth} & \textbf{Pose}  & \textbf{Synth} & \textbf{Pose} &  \\  
        [0.1em]
        \hline
        \hline
        $c_1$ & {75.47}  & {74.77}  & 74.47 & 74.94       & 75.24  & 74.78 & 75.90  & 74.95 &    \multirow{4}{*}{75.06}      \\  
        [0.1em]
        $c_{1,2}$  & {75.10}  & {74.70}  &  74.53 & 74.85  & 75.21 & 74.20 & 75.82 & 74.90 &             \\ 
        [0.1em]
        $c_{1-3}$     & 74.29  & 74.46   & 74.69 & 74.66 &  75.01  & {70.91}    & 74.59 & 75.55 & \\ 
        [0.1em]
        $c_{1-4}$  & {73.37}   & {74.33}  & 74.85 & 74.53       & 74.85 & 54.20 & 75.37 & 74.46& \\ 
        [0.1em]
        \hline
        \hline
        75\bm{$\%D_t$}  & \textbf{73.06}  & \textbf{73.49} & \textbf{73.46} &  \textbf{73.10} &  \textbf{73.92}  & \textbf{73.75} & \textbf{73.21} & \textbf{73.77} & 71.08     \\ 
        50\bm{$\%D_t$}  & \textbf{72.24}   & \textbf{71.02}   & \textbf{71.89}  & \textbf{72.08} & \textbf{73.01}   & \textbf{70.66}  & \textbf{72.53}     & \textbf{72.02} & 69.01      \\ 
        [0.1em]
        30\bm{$\%D_t$}  & \textbf{67.80}   & \textbf{64.81}    & \textbf{66.65}  & \textbf{66.20} & \textbf{67.42}      & \textbf{67.94} & \textbf{67.69} & \textbf{65.90} &  62.71     \\
        10\bm{$\%D_t$}  & \textbf{58.57}   & \textbf{59.66}   & \textbf{61.83}  & \textbf{59.79} & \textbf{60.50}   & \textbf{59.41}  & \textbf{61.93}     & \textbf{59.68} & 52.92      \\ 
        [0.1em]
        \hline
        \multicolumn{8}{c}{\vspace{-1mm}}\\
    \end{tabular}
    }
    \vspace{2mm}
    \resizebox{\columnwidth}{!}{
    \begin{tabular}{p{10mm} c c|c c|c c|c}
        \hline
        \multicolumn{8}{c}{\bm{$tCNN-IMU{D_s}^{LARa-Mb}$}} \\
        [0.2em]
        \textbf{Transf.} & \multicolumn{2}{c|}{D$_s$=JHMDB} & \multicolumn{2}{c}{D$_s$=CAD60} & \multicolumn{2}{c}{D$_s$=NTU RGB+D} & \textbf{Baseline} \\ 
        [0.1em]
        \textbf{layers} &  \textbf{Synth}  & \textbf{Pose}  & \textbf{Synth}  & \textbf{Pose}  & \textbf{Synth} & \textbf{Pose} &  \\  
        [0.1em]
        \hline
        \hline
        $c_1$ & {75.52}  & 76.04 & 75.36 & 75.32    & {75.05}  & {75.52} &    \multirow{4}{*}{75.09}      \\  
        [0.1em]
        $c_{1,2}$  & 75.23 & 75.69 & 74.99  & 75.02 &  {74.75} & 75.19&             \\ 
        [0.1em]
        $c_{1-3}$     & 75.02  & 75.11   & 74.32 & 74.69 &  {74.33} & 74.59 & \\ 
        [0.1em]
        $c_{1-4}$  & 74.82   & 74.86  & 73.99 & 74.36 &   {74.05} & 73.99 & \\ 
        [0.1em]
        \hline
        \hline
        75\bm{$\%D_t$}  & \textbf{74.82}  & \textbf{74.75} & 72.28 &   \textbf{73.65} & \textbf{73.92} & 73.33  & 73.51     \\ 
        50\bm{$\%D_t$}  & \textbf{71.26}   & \textbf{73.88}  &  \textbf{71.10} & \textbf{70.91} & \textbf{71.50}  & \textbf{70.91} & 70.30     \\ 
        [0.1em]
        30\bm{$\%D_t$}  & {66.38}   & {66.48}    & {67.01}  & \textbf{68.34} &  {67.45} &  {67.31} & 67.54    \\ 
        10\bm{$\%D_t$}  & \textbf{65.99}   & \textbf{65.18}    & \textbf{66.12}  & \textbf{66.95} &  \textbf{65.44} &  \textbf{62.80} & 57.47    \\
        \hline
        \multicolumn{6}{c}{\vspace{-10mm}}\\
    \end{tabular}
    }
\end{table}

Tab.~\ref{tab:TL_S_LARaOB} shows the performance of the transfer learning on the LARa-Mb as the target scenario. In this case, the performance of the four networks \textit{tCNN}$_{JHMDB}^{LARa-Mb}$, \textit{tCNN}$_{CAD60}^{LARa-Mb}$, \textit{tCNN}$_{Penn}^{LARa-Mb}$ and \textit{tCNN}$_{NTU}^{LARa-Mb}$ for both data sources, pose annotations and synthetic data, and different transferable layers remains similar to the \textit{tCNN}$^{LARa-Mb}$. The activities in the LARa-Mb dataset are performed in a warehouse environment, whereas the source datasets consider ADLs. The performance improves only when considering the $75\%$, $50\%$, $30\%$ or $10\%$ of the LARa-Mb. 

%% file: results/LaraMM.tex
There is no published research for HAR using LARa-MM. The baseline for LARa-MM is computed using a batch size of $200$ and $10$ epochs. However, the $\%wF1$ score for all the $D_s$ is comparable or significantly lower than the LARa-MM baseline, except for a few cases when the $D_t$ is less than $100\%$. These results constitute the first HAR performance for LARa-MM.
    
\begin{table}
    \caption{\scriptsize{Mean $wF1[\%]$ of the \textit{tCNN}$_{D_s}^{LARa-MM}$ using the joint poses and the synthetic data. The $N_{conv}$ changes from $c_1$ to $c_{1,2,3,4}$ keeping $100\%$ of the $D_t$. Subsequently, the $N_{conv}$ corresponding to highest $wF1[\%]$ is fixed and $\%D_t$ is varied. \textit{tCNN}$_{LARa-MM}$ is the baseline. Std $wF1[\%]$ lies around $0.01$. The performance of the \textit{tCNN-IMU}$_{D_s}^{MM}$ is similar, thus, not shown.}}
    \resizebox{\columnwidth}{!}{
    \begin{tabular}{p{10mm} c c|c c|c c|c c|c}
        \hline
        \multicolumn{10}{c}{\bm{$tCNN_{D_s}^{MM}$}} \\
        [0.2em]
        \textbf{Transf.} & \multicolumn{2}{c|}{D$_s$=JHMDB} & \multicolumn{2}{c|}{D$_s$=CAD60} & \multicolumn{2}{c|}{D$_s$=Penn} & \multicolumn{2}{c|}{D$_s$=NTU RGB+D} & \textbf{Baseline} \\ 
        [0.1em]
        \textbf{layers} &  \textbf{Synth}  & \textbf{Pose}  & \textbf{Synth}  & \textbf{Pose}  & \textbf{Synth} & \textbf{Pose}  & \textbf{Synth} & \textbf{Pose} &  \\
        [0.1em]
        \hline
        \hline
        $c_1$  & 65.14  & 64.72 & 65.01  & 65.67    & 64.27 & 65.97 & 64.33 & 64.79  & \multirow{4}{*}{65.71} \\ 
        [0.1em]
        $c_{1,2}$  & 65.21 & 64.12 & 64.80  & 65.05  & 63.73 & 64.81  & 64.09 & 63.21  &  \\
        [0.1em]
        $c_{1-3}$ & 65.45 & 62.04  & 64.56  & 60.26  & 63.19 & 62.35 & 63.51 & 63.11  &\\ 
        [0.1em]
        $c_{1-4}$  & 65.71 & 47.72 & 64.38 & 53.25 & 62.74 & 44.31 & 62.81 & 61.91  & \\ 
        [0.1em]
        \hline
        \hline
        75\bm{$\%D_t$} & \textbf{65.26} &   60.93 & 62.90  & 62.90 & 63.12   & 63.73 & 62.61 & 62.95  & 63.83 \\ 
        [0.1em]
        50\bm{$\%D_t$}  & \textbf{59.37} & {54.74} & 51.70  &  52.73 & \textbf{56.98} & 54.73 & {53.42} & \textbf{57.67}  & 53.14   \\ 
        [0.1em]
        30\bm{$\%D_t$}  & 53.80 &  50.09  & 45.75 & 52.20 & 49.90 & 45.88 & 51.12 & 50.84 & 51.19 \\ 
        10\bm{$\%D_t$}  & 47.41 & 46.81   & \textbf{51.36} & 48.95 & 49.01 & 46.59 & \textbf{51.57} & 43.32 & 47.06 \\ 
        \hline
        \multicolumn{6}{c}{\vspace{-3mm}}\\
    \end{tabular}
    }
    \label{tab:TL_S_LARaMM}
\end{table}

%% file: results/opportunity.tex
\input{Tables/locomotion_cnn}

Tab.~\ref{tab:TL_S_Loc} shows the classification performance of the transfer learning on the Opp-Loc dataset. The \textit{tCNN}$^{Loc}$ is considered as baseline. Only the \textit{tCNN}$_{JHMDB_{Pose}}^{Loc}$ using pose annotations and \textit{tCNN}$_{NTU_{}}^{Loc}$ improve the performance. In the case of having $[10,30,50]\%$ of the $D_t$, the pre-trained networks on JHMDB, CAD60 and Penn significantly improved the performance.
Tab.~\ref{tab:TL_S_Loc} also presents the performance of the three pre-trained architectures on the Opp-Ges $D_t$. In comparison with the Opp-Loc, pre-training the \textit{tCNN} with the three source datasets with both the pose annotations or the synthetic on-body devices influences the performance on the $D_t$ significantly, especially when transferring the first convolutional layer. The performance in all scenarios improves significantly when considering $50\%$ or $30\%$ of the $D_t$ and pre-training with the three $D_s$ using pose annotations or the synthetic on-body devices.

%% file: Tables/locomotion_cnn.tex
\begin{table}
    \caption {\scriptsize{Mean $wF1[\%]$ of the \textit{tCNN}$_{D_s}^{Loc}$ and \textit{tCNN}$_{D_s}^{Ges}$ using the joint poses and the synthetic data. The $N_{conv}$ changes from $c_1$ to $c_{1,2,3,4}$ keeping $100\%$ of the $D_t$. Subsequently, the $N_{conv}$ corresponding to highest $wF1[\%]$ is fixed and $\%D_t$ is varied. The $tCNN_{Loc}$, trained on the training set of the Opp-Loc and -Ges, denotes the baseline. Std $wF1[\%]$ lies around $0.01$. The performance of the \textit{tCNN-IMU}$_{D_s}^{Loc}$ and \textit{tCNN-IMU}$_{D_s}^{Ges}$ is similar, thus, not shown.}}
    \resizebox{\columnwidth}{!}{
    \begin{tabular}{p{10mm} c c|c c|c c|c c|c}
        \hline
        \multicolumn{10}{c}{\bm{$tCNN_{D_s}^{Loc}$}} \\
        [0.2em]
        \textbf{Transf.} & \multicolumn{2}{c|}{D$_s$=JHMDB} & \multicolumn{2}{c|}{D$_s$=CAD60} & \multicolumn{2}{c|}{D$_s$=Penn} & \multicolumn{2}{c|}{D$_s$=NTU RGB+D} & \textbf{Baseline} \\ 
        [0.1em]
        \textbf{layers} &  \textbf{Synth}  & \textbf{Pose}  & \textbf{Synth}  & \textbf{Pose}  & \textbf{Synth} & \textbf{Pose}  & \textbf{Synth} & \textbf{Pose} &  \\
        [0.1em]
        \hline
        \hline
        $c_1$ & \textbf{86.70}   & \textbf{86.84} & \textbf{86.87} & 86.28  & \textbf{86.70}  & 86.10 & 86.34 & \textbf{86.51} & \multirow{4}{*}{85.65 }     \\ 
        $c_{1,2}$& {86.54} & \textbf{86.80} & 86.40 & 86.15  & {86.54} & 85.97 &   \textbf{86.62} & \textbf{86.77} &   \\ 
        $c_{1-3}$ & 86.40 & \textbf{86.74} & 86.21 & 86.08  & 86.20& 85.86 &  \textbf{87.01} & \textbf{86.96} &  \\ 
        $c_{1-4}$ & 86.26 & \textbf{86.72} & 86.02 & 85.96 & 86.01 & 85.49 &  \textbf{87.21} & \textbf{87.05} &   \\
        \hline
        \hline
        75\bm{$\%D_t$} & 85.90  & \textbf{87.33} & 86.01 & 85.67& 85.66 & 85.13 &  \textbf{86.67} & 86.26 & 85.67       \\ 
        50\bm{$\%D_t$} & {85.18}  & \textbf{85.60} & \textbf{85.62}   & \textbf{85.46}  & 85.10  & 84.13 & \textbf{85.62}   & 85.03  & 84.44     \\ 
        30\bm{$\%D_t$} & \textbf{84.23} & {84.05} & \textbf{84.32} & 83.74  & \textbf{84.20} & 83.83 &  83.78  & 81.46 & 83.11     \\ 
        10\bm{$\%D_t$} & \textbf{55.16} & \textbf{54.77} & \textbf{54.90} & 53.12  & \textbf{55.33} & \textbf{55.54} & 53.90 & 53.61  & 52.60      \\       
        \hline
        \multicolumn{6}{c}{\vspace{-1mm}}\\
    \end{tabular}
}
    \resizebox{\columnwidth}{!}{
    \begin{tabular}{p{10mm} c c|c c|c c|c c|c}
        \hline
        \multicolumn{10}{c}{\bm{$tCNN_{D_s}^{Ges}$}} \\
        [0.2em]
        \textbf{Transf.} & \multicolumn{2}{c|}{D$_s$=JHMDB} & \multicolumn{2}{c|}{D$_s$=CAD60} & \multicolumn{2}{c|}{D$_s$=Penn}  & 
        \multicolumn{2}{c|}{D$_s$=NTU}  & 
        \textbf{Baseline} \\ 
        [0.1em]
        \textbf{layers} &  \textbf{Synth}  & \textbf{Pose}  & \textbf{Synth}  & \textbf{Pose}  & \textbf{Synth} & \textbf{Pose}  & \textbf{Synth} & \textbf{Pose} &  \\ 
        [0.1em]
        \hline
        \hline
        $c_{1}$     & \textbf{88.66} & \textbf{89.05} &\textbf{88.82}  & \textbf{88.72}  &      \textbf{88.77}   &  \textbf{88.96}   &  \textbf{89.19} &       \textbf{89.03} & \multirow{4}{*}{87.36}     \\  
        $c_{1,2}$  & \textbf{88.83} & \textbf{88.97}  & \textbf{88.75}  & \textbf{88.80} & \textbf{88.67} & \textbf{88.53} &  \textbf{88.93} &  \textbf{88.81}& \\  
        $c_{1-3}$  &  \textbf{88.79} &    \textbf{88.94}   &  \textbf{88.92} &     \textbf{88.99}    & \textbf{88.58}     &      \textbf{88.12} &  \textbf{88.63} &  \textbf{88.59} &\\  
        $c_{1-4}$ &  \textbf{88.87} &  \textbf{88.75}  &  \textbf{88.99}  &  \textbf{89.11}  & \textbf{88.51}     &  \textbf{87.84} &  \textbf{88.46} &      \textbf{88.56}& \\ 
        \hline
        \hline
        75\bm{$\%D_t$}  & {87.33} & {87.41} & \textbf{87.44}  & {87.12}  & \textbf{87.92}   &   {87.21}    & 
        \textbf{88.14} &
        \textbf{88.28} & 86.40                        \\ 
        50\bm{$\%D_t$}  & \textbf{87.06} & \textbf{87.04}  & \textbf{87.01}  & \textbf{86.77}  & \textbf{87.34}   &   \textbf{87.52}   & \textbf{86.96} & \textbf{86.77} & 85.40                        \\ 
        30\bm{$\%D_t$}  & \textbf{85.22} & \textbf{85.06}  & \textbf{84.62} & \textbf{84.93}  & \textbf{86.70}    &   \textbf{85.02}    &
        84.80 & 84.60 & 83.40                        \\ 
        10\bm{$\%D_t$}  & {76.47} & {75.99}  & {76.54} & {76.11}  &     75.67     &  75.67 & 75.67 & 75.67  & 75.57  \\
        \hline
        \multicolumn{6}{c}{\vspace{-7mm}}\\
    \end{tabular}
    }
    \label{tab:TL_S_Loc}
\end{table}

%% file: chapters/05_conclusions.tex
\section{Conclusion}
\label{sec:conclusion}

This paper proposes to use pose annotations from video datasets as an input stream for solving multichannel-time series human activity recognition (HAR) using transfer learning. Four different datasets comprising ground truth pose estimation from videos are deployed as the $D_s$. The sequences of pixel coordinates of human joints are considered as simplifications of pose estimations. A temporal CNN that processes inertial measurements per channel with late fusion is trained, considering the pixel coordinates as individual channels. The learned temporal convolutional layers are used to initialize architectures on three benchmark datasets for HAR. Experiments regarding the number of transferred layers, the two versions of the $D_s$, and a proportion of the $D_t$ are carried out. In general, when transferring only the first convolutional layer, the performance showed a positive influence on the task, independently of the data source. Transfer learning helps when a small proportion of the training target set is available. Besides, source and target datasets with shared activities showed some improvements. These findings suggest that simple local temporal relations are rather generic, thus transferable. The more complex and task-related filters are not transferable. Experimentation using the networks, such as \textit{RNNs} or Fully-convolutional \textit{CNN}, could further improve the results.